\documentclass[conference]{IEEEtran}
\IEEEoverridecommandlockouts

\usepackage{cite}
\usepackage{amsmath,amssymb,amsfonts}
\usepackage{algorithmic}
\usepackage{graphicx}
\usepackage{textcomp}
\usepackage{xcolor}
\usepackage{url}
\usepackage{booktabs}
\usepackage{lmodern}
\usepackage{listings}

\lstdefinelanguage{DSL}{
  basicstyle=\fontsize{6pt}{7pt}\selectfont\ttfamily,
  breaklines=true,
  frame=none,
  aboveskip=2pt,
  belowskip=2pt,
  showstringspaces=false
}

\def\BibTeX{{\rm B\kern-.05em{\sc i\kern-.025em b}\kern-.08em
    T\kern-.1667em\lower.7ex\hbox{E}\kern-.125emX}}

\begin{document}

\title{Bridging the Sim-to-Real Gap in Semiconductor Visual Program Synthesis via Input Binarization}

\author{
\IEEEauthorblockN{Yusuke Ohtsubo\textsuperscript{1}, Kota Dohi\textsuperscript{1}, Koichiro Yawata\textsuperscript{1}, Koki Takeshita\textsuperscript{1}, Tatsuya Sasaki\textsuperscript{1}}
\IEEEauthorblockA{\textsuperscript{1}Hitachi, Ltd. R\&D Group, Tokyo, Japan \\
\{yusuke.ohtsubo.nb, kota.dohi.gr, koichiro.yawata.rt, koki.takeshita.cx, tatsuya.sasaki.gp\}@hitachi.com}
}

\maketitle
\let\oldthefootnote\thefootnote
\let\thefootnote\relax\footnotetext{Code is available at \url{https://github.com/YusukeO/eusipco2026-sem-dsl}.}
\let\thefootnote\oldthefootnote

\begin{abstract}
Precise parametric control over circuit geometry is essential for semiconductor inspection, yet obtaining sufficient real training data remains costly. Although generative models such as diffusion models and Generative Adversarial Networks (GANs) can augment training data, they cannot guarantee the nanometer-scale geometric accuracy required for metrology tasks. We propose a visual program synthesis framework in which a Vision-Language Model (VLM) converts inspection images into editable Domain-Specific Language (DSL) code describing circuit geometries, enabling controlled generation of training data with exact parameter manipulation. Because the VLM is trained solely on synthetic DSL-rendered data, a domain gap arises when processing real Scanning Electron Microscope (SEM) images. We bridge this gap with an input binarization strategy that strips SEM-specific texture and noise, letting the model focus on geometric structure. On the MIIC dataset, binarized inputs improve the mean Dice coefficient from 0.4393 to 0.5256 over the raw-input baseline, demonstrating that simple texture abstraction substantially mitigates the sim-to-real gap.
\end{abstract}

\begin{IEEEkeywords}
Semiconductor inspection, domain-specific language, vision-language model, visual program synthesis, binarization
\end{IEEEkeywords}

\section{Introduction}

\begin{figure}[!t]
\centerline{\includegraphics[width=0.9\linewidth]{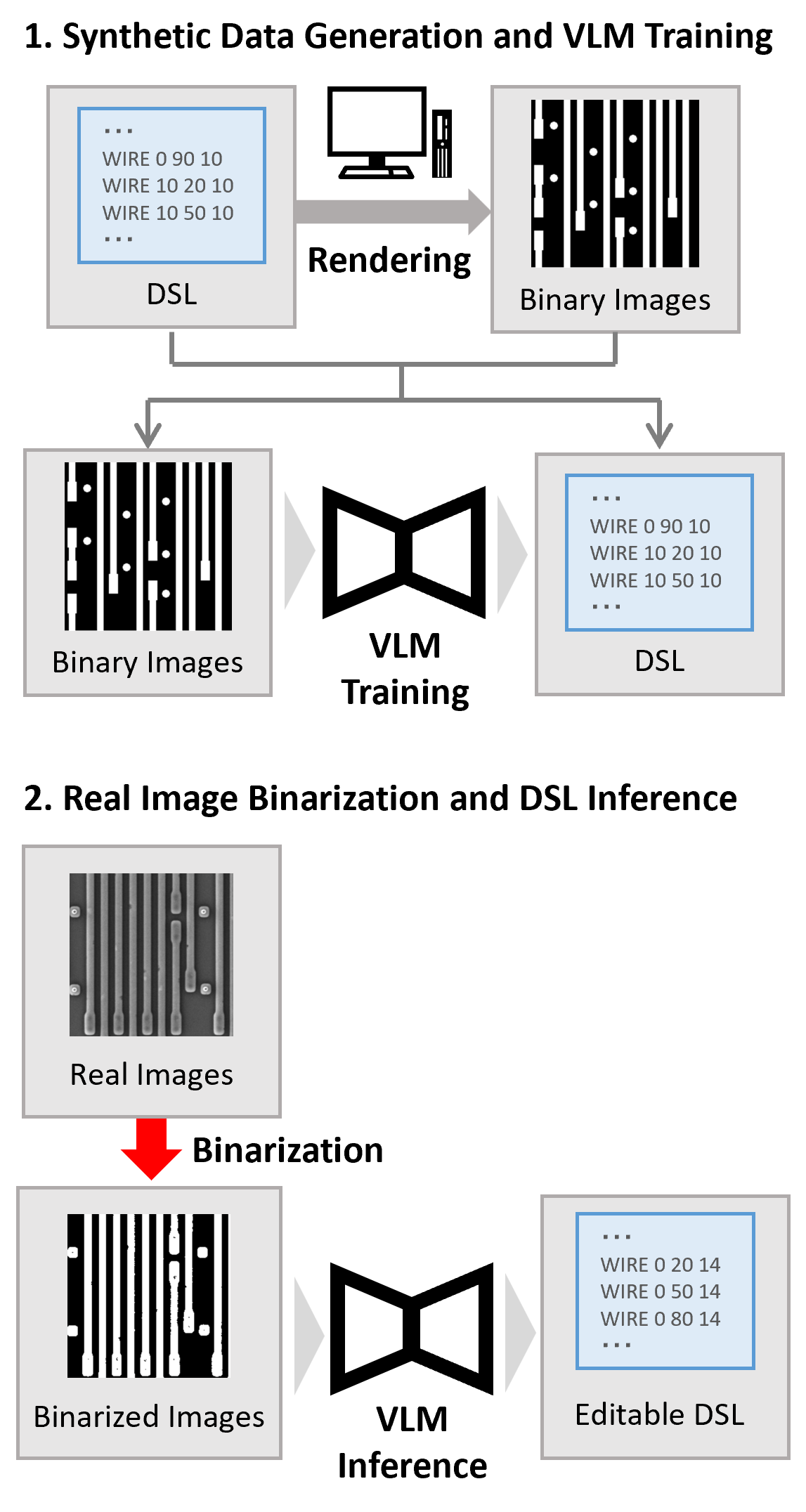}}
\caption{Overview of the proposed visual program synthesis framework. (1) The VLM is trained on synthetic data pairs created via DSL rendering. (2) During inference, real SEM images are binarized to bridge the sim-to-real gap, enabling the robust reconstruction of editable DSL.}
\label{fig:overview}
\end{figure}

We propose a visual program synthesis framework that converts semiconductor inspection images into a reproducible domain-specific language (DSL).
DSLs are purpose-built languages that trade generality for concise, precise, and editable domain representations. With direct parameter control, built-in rule checking, and high reproducibility, they are standard in semiconductor design.
By manipulating code parameters such as linewidth and length, we can actively generate training data for downstream tasks, including circuit linewidth measurement and denoising. This also enables controlled generation of hard negatives, critical areas, and topology-induced domain shifts that are difficult to collect from real samples.

Accurate manipulation at the nanometer scale is essential for semiconductor circuit images. While generative models like diffusion models~\cite{rombach2022ldm} and Generative Adversarial Networks~\cite{goodfellow2014gan} offer natural-language-based control, they are ill-suited for such precise geometric control. In contrast, code-level description allows for operations directly linked to design parameters, making it highly suitable for this domain.

Existing research on image-to-code conversion, such as ChartCoder~\cite{zhao2025chartcoder} and pix2code~\cite{beltramelli2018pix2code}, is limited to environments where image-code pairs are readily available (e.g., web data). Other approaches like Im2Vec~\cite{reddy2021im2vec} convert general images into primitive sets but require specialized network designs for each domain. Unlike rigid Computer-Aided Design (CAD) recovery tools with fixed syntax, our approach treats the language specification itself as a design variable. The same Vision-Language Model (VLM) architecture can therefore be reused across inspection domains by changing only the Domain-Specific Language (DSL).

We define a DSL tailored to our objectives and train a VLM using a large volume of synthetic data generated via this DSL. However, a domain gap exists between the DSL-derived synthetic data and the real SEM images used during inference, primarily due to noise and texture inherent to SEM imaging. To address this, we investigate the following research question: Does inference-time input binarization reduce the domain gap when real SEM images are input to the VLM?
It is not self-evident that binarization alone suffices, as other geometric domain gaps (e.g., corner rounding, beam-induced distortion) also exist. However, we hypothesize that texture and noise are the dominant factors hindering VLM inference, and thus texture abstraction via binarization yields substantial gains.
Furthermore, this aligns with downstream workflows like Design Rule Checking and connectivity analysis, where binarization is a standard precursor.
The main contribution of this study is demonstrating that input binarization effectively reduces structural domain gaps by abstracting texture information in real-world applications.

\section{Related Work}

\subsection{Parametric Generation and Image-to-Code}
Parametric or code-based image generation is attractive when geometric precision matters. Prior work recovers CAD commands from drawings~\cite{qin2025drawing2cad} or generates executable CAD scripts from text~\cite{xie2025text}. However, these methods often rely on predefined, rigid command sets typical of standard CAD software. We extend this paradigm to semiconductor inspection by introducing a flexible framework where the language itself is a design variable. In this study, we define a lightweight DSL for Manhattan layouts and learn image-to-DSL translation with a VLM. We further verify the effect of input binarization in this context, which may also offer insights for CAD recovery tasks.

\subsection{Structure-aware Domain Adaptation}
In document image analysis, binarization is established as a method to remove noise and extract domain-invariant structural features from degraded images~\cite{tensmeyer2017binarization}.
Our proposed "binary input" extends these structure-centric adaptation strategies. By intentionally discarding SEM-specific noise and texture, we create an environment where the VLM can focus solely on the geometric features necessary for DSL inference.

\section{Methodology}

This section describes the definition of the DSL for semiconductor geometries, synthetic data generation, VLM training, and the binarization preprocessing for real images.

\subsection{Design of Semiconductor DSL}
\begin{figure}[t]
\centering
\renewcommand{\arraystretch}{1.1}
\setlength{\tabcolsep}{1pt}

\begin{tabular}{cc}
% --- (a) Manhattan (12 lines) ---
\begin{minipage}[c]{0.65\linewidth}
\textbf{(a) Manhattan Wiring Pattern} \\
\vspace{-2pt}
\begin{lstlisting}[language=DSL]
CANVAS 256 256
WIRE 0 12 8  0 0  0 0  H 256
WIRE 0 28 8  0 0  20 12  H 80
WIRE 100 28 8  20 12  0 0  H 156
WIRE 0 44 8  0 0  0 0  H 180
WIRE 200 44 8  20 12  0 0  H 56
... (10 lines omitted) ...
WIRE 160 220 8  20 12  20 12  H 60
WIRE 0 236 8  0 0  0 0  H 256
\end{lstlisting}
\end{minipage} &
\begin{minipage}[c]{0.31\linewidth}
\centering
\includegraphics[width=0.98\linewidth]{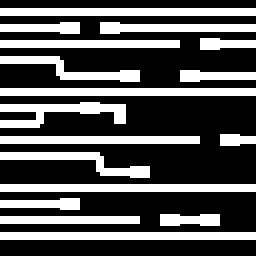}
\end{minipage} \\
\midrule

% --- (b) Horizontal (12 lines) ---
\begin{minipage}[c]{0.65\linewidth}
  \vspace{4pt}
  \textbf{(b) Horizontal Line Pattern} \\
  \vspace{-2pt}
\begin{lstlisting}[language=DSL]
CANVAS 256 256
WIRE 0 20 14  0 0  0 0  H 256
WIRE 0 50 14  0 0  0 0  H 256
WIRE 0 80 14  0 0  0 0  H 256
WIRE 0 110 14  0 0  0 0  H 256
WIRE 0 140 12  0 0  0 0  H 256
... (4 lines omitted) ...
WIRE 0 235 12  0 0  0 0  H 256
WIRE 0 245 10  0 0  0 0  H 256
\end{lstlisting}
\end{minipage} &
\begin{minipage}[c]{0.31\linewidth}
\centering
\includegraphics[width=0.98\linewidth]{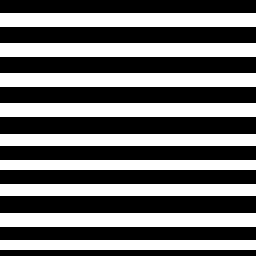}
\end{minipage} \\
\midrule

% --- (c) Vertical (12 lines) ---
\begin{minipage}[c]{0.65\linewidth}
  \vspace{4pt}
  \textbf{(c) Vertical Stripe Pattern} \\
  \vspace{-2pt}
\begin{lstlisting}[language=DSL]
CANVAS 256 256
WIRE 12 0 10  0 0  30 14  V 60
WIRE 12 100 10  30 14  30 14  V 80
VIA 35 40 6
WIRE 55 0 10  0 0  0 0  V 256
WIRE 75 0 10  0 0  30 14  V 200
... (9 lines omitted) ...
WIRE 215 0 10  0 0  30 14  V 180
WIRE 235 0 10  0 0  0 0  V 256
\end{lstlisting}
\end{minipage} &
\begin{minipage}[c]{0.31\linewidth}
\centering
\includegraphics[width=0.98\linewidth]{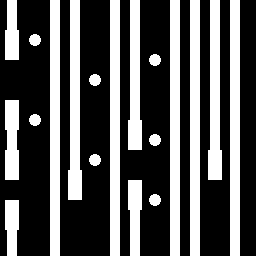}
\end{minipage} \\

\end{tabular}
\caption{Examples of the proposed DSL and corresponding rendered images. (a) Manhattan wiring, (b) Horizontal lines, and (c) Vertical stripes.}
\label{fig:dsl_examples}
\end{figure}

We designed a DSL tailored for Manhattan wiring architectures.
Unlike raster images, this DSL describes circuit patterns using vector-based geometric commands.
As shown in Fig.~\ref{fig:dsl_examples}, the DSL structure follows a hierarchical definition starting with the canvas, followed by wiring paths and vias.

First, the drawing area is defined. The coordinate system originates at the top-left $(0,0)$, with the x-axis extending right and the y-axis down:
\begin{equation}
    \texttt{CANVAS}(W, H),
\end{equation}
where $W$ and $H$ denote the width and height of the image (e.g., $256 \times 256$ in Fig.~\ref{fig:dsl_examples}).

The core component, \texttt{WIRE}, abstracts a continuous conductive trace.
To capture semiconductor-specific geometries such as contact pads or lithography artifacts, we explicitly parameterize the "dogbone" shape---a local widening at wire ends.
The command is defined as:
\begin{equation}
    \texttt{WIRE}(x_0, y_0, w_{base}, l_{s}, w_{s}, l_{e}, w_{e}, \mathcal{S}).
    \label{eq:WIRE_def}
\end{equation}
Here, $(x_0, y_0)$ is the starting coordinate and $w_{base}$ is the primary linewidth.
The parameters $l_{s}, w_{s}$ (length, width) and $l_{e}, w_{e}$ define the geometry of the start and end caps, respectively.
For instance, setting $w_{s} > w_{base}$ creates a dogbone shape typically used for stable via connections, as observed in the vertical lines of Fig.~\ref{fig:dsl_examples}(c).
If $l_{s}=0$, the wire has a standard flush end.

$\mathcal{S}$ represents the routing path as a sequence of Manhattan segments.
Instead of absolute coordinates, $\mathcal{S}$ is described as a chain of relative movements:
\begin{equation}
    \mathcal{S} = \{ (d_1, l_1), (d_2, l_2), \dots \},
\end{equation}
where $d_i \in \{H, V\}$ indicates the direction (Horizontal/Vertical) and $l_i$ specifies the length.
This corresponds to the notation like \texttt{H 60 V 16} in Fig.~\ref{fig:dsl_examples}(a), enabling flexible "one-stroke" routing descriptions.

Finally, interlayer connections are represented by:
\begin{equation}
    \texttt{VIA}(x, y, r),
\end{equation}
which places a circular via of radius $r$ at $(x, y)$.
By combining these primitives, the DSL can compactly represent complex circuit topologies ranging from simple lines to maze-like patterns.

\subsection{Synthetic Data Generation for VLM Training}
\label{sec:synthetic_gen}
We categorized real SEM image topologies into three types: \textit{Vertical Stripes}, \textit{Horizontal Lines}, and \textit{Manhattan}. To prevent the model from overfitting to specific spatial frequencies or densities, we treated DSL structural parameters as broad probability distributions rather than fixed values, thereby maximizing scale and structural diversity.

Specifically, we randomized grid divisions and wiring densities for each image generation, where $\mathcal{U}$ denotes the uniform distribution. (1) For \textit{Vertical Stripes}, we varied the number of vertical columns $N_{col} \sim \mathcal{U}(8, 16)$ and wiring probability $p_{wire} \sim \mathcal{U}(0.5, 0.9)$. (2) For \textit{Horizontal Lines}, we manipulated the number of bands $N_{band} \sim \mathcal{U}(8, 16)$ and gap rate $p_{gap} \sim \mathcal{U}(0.0, 0.4)$. (3) For \textit{Manhattan}, we varied lane counts $N_{lane} \sim \mathcal{U}(12, 24)$ and bend probability $p_{bend} \sim \mathcal{U}(0.4, 0.8)$ to cover structures ranging from simple wires to complex maze-like patterns.
These ranges were determined based on statistical analysis of the MIIC dataset to ensure the generated patterns cover real-world circuit geometries.
Across all three types, the linewidth $w_{base}$ was sampled from a uniform distribution of $7$--$15$\,px, and one-sided salt impulse noise was injected independently by setting each pixel to the salt value with probability 0.01. Because the images are binary and the corruption is impulse-like, we report this perturbation by its pixel corruption rate rather than a signal-to-noise ratio. Through this pipeline, we generated 19,200 image-DSL pairs (Train: 18,900, Eval: 300) for VLM training.

\subsection{VLM Architecture and Training}
We adopted Qwen3-VL-8B~\cite{bai2025qwen3vl} as our model and performed supervised fine-tuning (SFT) to translate input images into DSL.
Let $\mathcal{D}_{\text{syn}} = \{(x^{\text{syn}}_n, \mathbf{t}_n)\}_{n=1}^{N}$ denote the synthetic training dataset generated from the DSL (Section~\ref{sec:synthetic_gen}), where $x^{\text{syn}}_n$ is the rendered image and $\mathbf{t}_n = (t_{n,1}, t_{n,2}, \dots, t_{n,L_n})$ is the corresponding ground-truth DSL sequence of length $L_n$.
Note that $x^{\text{syn}}_n$ is binary by construction (the renderer outputs black/white patterns with optional salt noise), thus no additional binarization is applied during training.

The optimization objective is to find the optimal parameters $\theta^*$ that minimize the negative log-likelihood of the conditional probability for the next token prediction:
\begin{equation}
\theta^* = \operatorname*{argmin}_\theta \left( - \sum_{n=1}^{N} \sum_{i=1}^{L_n} \log P_\theta(t_{n,i} \mid x^{\text{syn}}_n, t_{n,<i}) \right),
\label{eq:sft_obj}
\end{equation}
where $t_{n,<i}$ represents the autoregressive context.
All parameters, including the vision encoder and multimodal projector, were trainable. The main hyperparameters were configured as follows: batch size 8, gradient accumulation steps 12, learning rate 2.0e-5 (using a cosine scheduler with a 0.1 warmup ratio), training epochs 3.0, weight decay 0.01, and max gradient norm 1.0.

\subsection{Structure-aware Input Binarization}
While synthetic images generated by the DSL are binary by construction, real SEM images contain high-frequency noise, material contrast textures, and edge effects characteristic of electron microscopy.
To bridge this sim-to-real gap at inference time, we transform a grayscale real image $x^{\text{real}}$ into a black-and-white binary image $\mathcal{B}(x^{\text{real}})$ using global thresholding before feeding it to the VLM (see Fig.~\ref{fig:qualitative_comparison}).
This preprocessing aligns real inputs with the binary distribution of the synthetic training images, thereby reducing texture-induced domain mismatch.
Global-threshold binarization is also a common segmentation step in Scanning Electron Microscope (SEM) image processing pipelines~\cite{Fraczek2006EUSIPCO}. The threshold was empirically fixed at 100 based on visual inspection of a few representative samples and was not tuned on the MIIC test set; lower values can retain SEM texture as spurious structures, whereas higher values can remove thin wires or small vias.
This step serves as geometric normalization because the Domain-Specific Language encodes binary regions and boundaries rather than grayscale texture.
\section{Experiments}

We verified whether the proposed method (binary input) outperforms the baseline (raw input) in structural reconstruction of real images. We used the aforementioned synthetic data for training and the test set (1034 normal data images) of the MIIC dataset~\cite{huang2021miic} for evaluation. Binarized real images served as the ground truth (GT).
We compared two conditions: the proposed binary input, where real images are binarized at a threshold of 100, and the baseline raw input, which uses grayscale real images.

\subsection{Evaluation Metrics}
To validate the generated DSL, we compared rendered images with the ground truth (GT) using region-, boundary-, and topology-level metrics. For region-level consistency, we used Intersection over Union (IoU) and Dice coefficient. For geometric fidelity, we used boundary F1 score (BF1)~\cite{csurka2013bfscore} and average symmetric surface distance (ASSD), as boundary accuracy is important in high-resolution Scanning Electron Microscope-based integrated circuit analysis~\cite{Pollach2022EUSIPCO}. For topology, we used skeleton F1 score (SkF1)~\cite{bastani2018roadtracer}, which is sensitive to breaks and shorts compared with pixel-based IoU.

\section{Results and Discussion}
\subsection{Sim-to-Real Reconstruction Performance}
Executable outputs were obtained for 957 binary-input images (97.1\%) and 1019 raw-input images (98.5\%). Table~\ref{tab:quantitative_comparison} shows the quantitative evaluation of rendered executable DSL outputs. Binary input improved all retained metrics; Dice coefficient increased by 0.0863, from $0.4393 \pm 0.0980$ to $0.5256 \pm 0.0912$.

\begin{table}[htbp]
\caption{MIIC reconstruction performance. Values are mean $\pm$ standard deviation over executable outputs.}
\begin{center}
\begin{tabular}{lcc}
\toprule
\textbf{Metric} & \textbf{Raw} & \textbf{Binary} \\
\midrule
IoU & 0.2865 $\pm$ 0.0802 & \textbf{0.3619 $\pm$ 0.0882} \\
Dice coefficient & 0.4393 $\pm$ 0.0980 & \textbf{0.5256 $\pm$ 0.0912} \\
BF1@2px & 0.4054 $\pm$ 0.1106 & \textbf{0.4412 $\pm$ 0.1098} \\
SkF1@1px & 0.1276 $\pm$ 0.0960 & \textbf{0.1746 $\pm$ 0.1145} \\
ASSD & 4.7768 $\pm$ 3.4596 & \textbf{4.1327 $\pm$ 1.2757} \\
\bottomrule
\end{tabular}
\label{tab:quantitative_comparison}
\end{center}
\end{table}

Next, to assess the impact of the domain gap, we measured the model's upper performance limit using the 300 synthetic evaluation samples described in Section~\ref{sec:synthetic_gen}, which are identical to the training distribution (Table~\ref{tab:synthetic_performance}).

\begin{table}[htbp]
\caption{Synthetic-data performance without domain gap.}
\begin{center}
\begin{tabular}{lc}
\toprule
\textbf{Metric} & \textbf{Score} \\
\midrule
IoU & 0.4760 \\
Dice coefficient & 0.6340 \\
BF1@2px & 0.5313 \\
SkF1@1px & 0.3055 \\
ASSD & 3.4271 \\
\bottomrule
\end{tabular}
\label{tab:synthetic_performance}
\end{center}
\end{table}

While binary input improved performance on real images, scores remained inferior to those on synthetic data. This suggests that while binarization reduced texture/noise-related gaps, other non-color domain gaps, such as subtle shape differences, persist.

\subsection{Qualitative Analysis}
\begin{figure}[htbp]
\centering
\setlength{\tabcolsep}{1pt}
\begin{tabular}{cccc}
\scriptsize Input (raw) & \scriptsize Input (binary) & \scriptsize Output (raw) & \scriptsize Output (binary) \\
\includegraphics[width=0.24\linewidth]{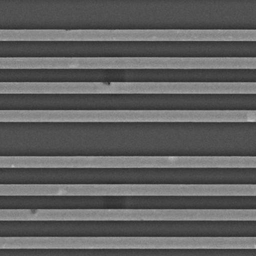} &
\includegraphics[width=0.24\linewidth]{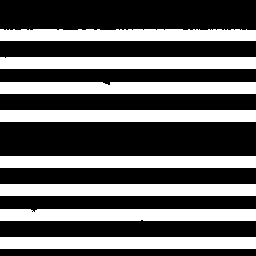} &
\includegraphics[width=0.24\linewidth]{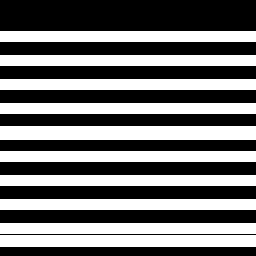} &
\includegraphics[width=0.24\linewidth]{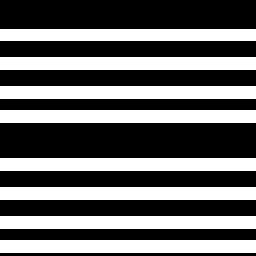} \\
\includegraphics[width=0.24\linewidth]{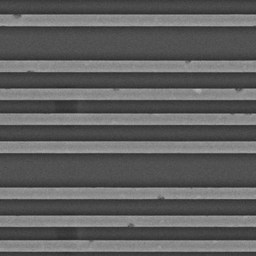} &
\includegraphics[width=0.24\linewidth]{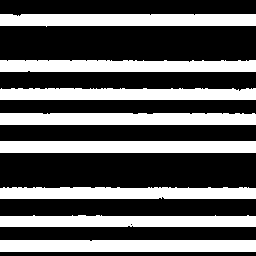} &
\includegraphics[width=0.24\linewidth]{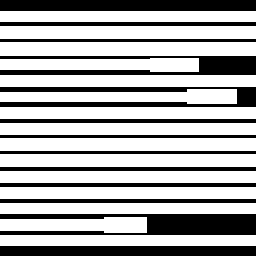} &
\includegraphics[width=0.24\linewidth]{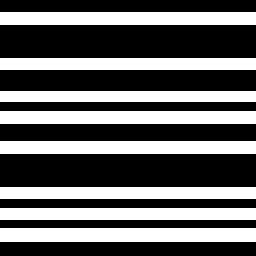} \\
\includegraphics[width=0.24\linewidth]{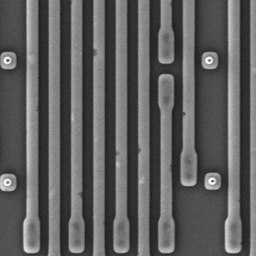} &
\includegraphics[width=0.24\linewidth]{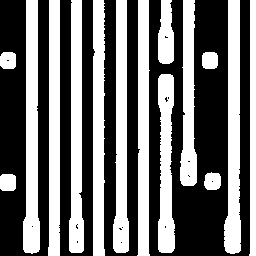} &
\includegraphics[width=0.24\linewidth]{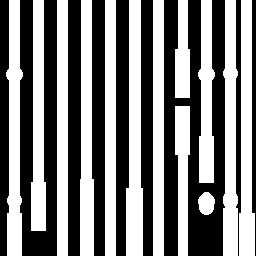} &
\includegraphics[width=0.24\linewidth]{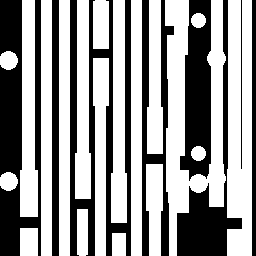} \\
\includegraphics[width=0.24\linewidth]{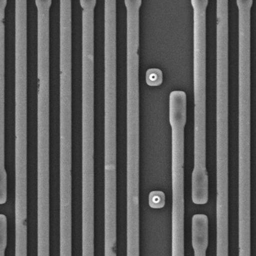} &
\includegraphics[width=0.24\linewidth]{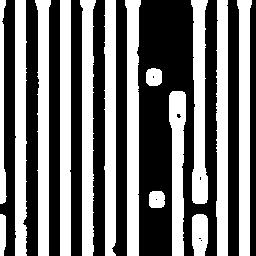} &
\includegraphics[width=0.24\linewidth]{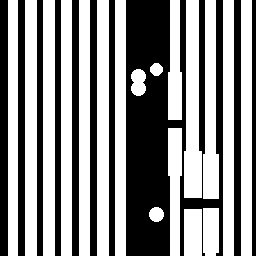} &
\includegraphics[width=0.24\linewidth]{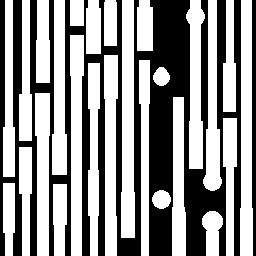} \\
\includegraphics[width=0.24\linewidth]{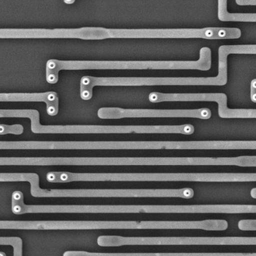} &
\includegraphics[width=0.24\linewidth]{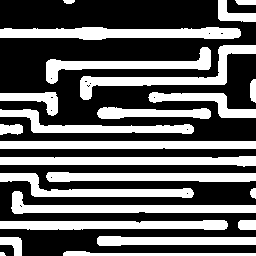} &
\includegraphics[width=0.24\linewidth]{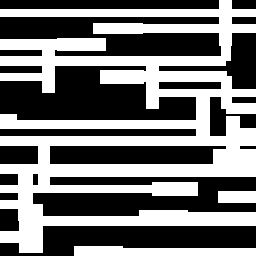} &
\includegraphics[width=0.24\linewidth]{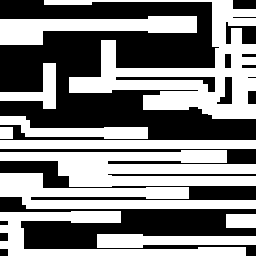} \\
\includegraphics[width=0.24\linewidth]{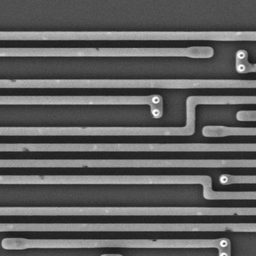} &
\includegraphics[width=0.24\linewidth]{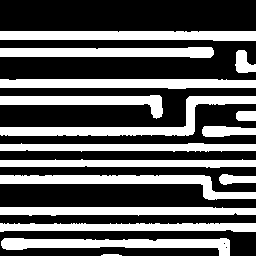} &
\includegraphics[width=0.24\linewidth]{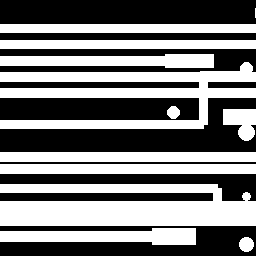} &
\includegraphics[width=0.24\linewidth]{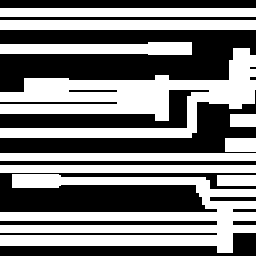} \\
\end{tabular}
\caption{Qualitative comparison of reconstruction results. Columns show (from left to right): input raw image, input binary image, output from raw input, and output from binary input.
Simple structures (e.g., horizontal line patterns in the top rows) show high reconstruction accuracy, whereas the accuracy decreases as the structures become more complex (towards the bottom rows).}
\label{fig:qualitative_comparison}
\end{figure}

Fig.~\ref{fig:qualitative_comparison} demonstrates that the proposed method reconstructs circuit patterns more accurately than the raw input baseline. In the second horizontal-line example, the reconstruction from raw input differs markedly from the input because weak grayscale texture and edge contrast are interpreted as additional geometry, resulting in over-generated thin wires. Reconstruction fidelity still declines as circuit patterns become more intricate, even with binarized input, suggesting a trade-off between pattern complexity and accuracy, which we analyze next.

\subsection{Impact of Structural Complexity}
\begin{figure}[htbp]
\centerline{\includegraphics[width=\linewidth]{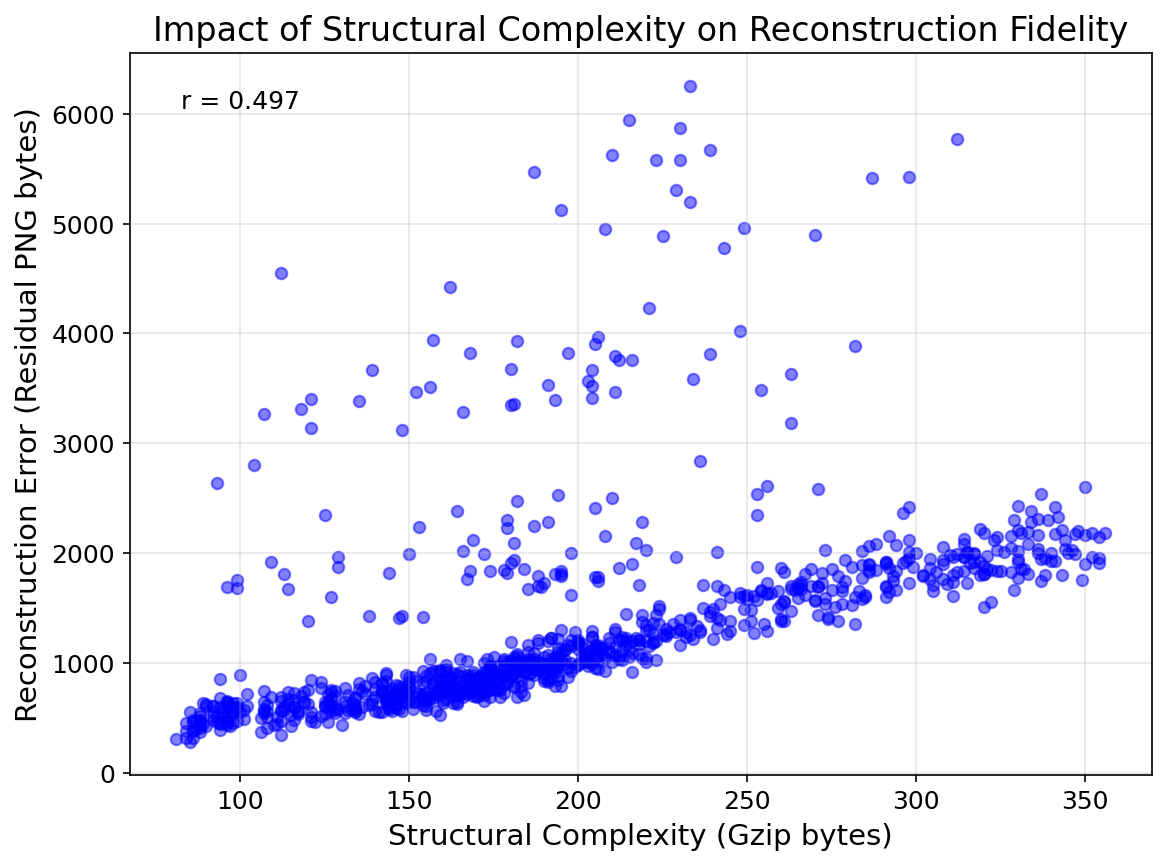}}
\caption{Impact of structural complexity on reconstruction fidelity for binary input. The x-axis approximates geometric information content by compressed Domain-Specific Language size, and the y-axis approximates unexplained residual entropy by Portable Network Graphics file size.}
\label{fig:dsl_length_vs_error}
\end{figure}

To quantitatively investigate the trade-off between structural complexity and reconstruction accuracy, we defined two metrics for the axes in Fig.~\ref{fig:dsl_length_vs_error}.
First, we quantified Structural Complexity using the byte size of Gzip-compressed DSL. Unlike simple character counts, compression eliminates redundancies (e.g., whitespace and boilerplate), providing a more accurate measure of the substantial geometric information content.
Second, we measured Reconstruction Error using the file size of the residual image (exclusive OR between ground truth and reconstruction) saved as Portable Network Graphics (PNG). This metric captures the entropy of unexplained information, making it more suitable than pixel-wise counts.
The analysis revealed a positive correlation: as DSL description length increases, reconstruction error grows (Fig.~\ref{fig:dsl_length_vs_error}).

The near-linear trend in Fig.~\ref{fig:dsl_length_vs_error} is expected because both axes are information-content proxies: denser layouts generally require longer compressed Domain-Specific Language descriptions and leave higher-entropy residuals after reconstruction. The outliers mainly correspond to topology-level failures, such as missing, broken, or hallucinated wires and vias. These failures may arise from insufficient feature extraction by the vision encoder or limited long-sequence generation by the language-model decoder; if the latter is dominant, a more concise Domain-Specific Language would be effective.

\section{Conclusion}
We proposed a visual program synthesis framework that converts semiconductor inspection images into editable DSL, facilitating active training data generation.
Our primary contribution is demonstrating that input binarization effectively bridges the domain gap in semiconductor inspection tasks by abstracting texture and extracting only geometric structures.
This substantially reduced the sim-to-real gap and improved structural reconstruction accuracy compared with raw image input.

By ensuring structural fidelity on real images, our framework establishes a reliable foundation for the intended application: procedural data augmentation.
Future work will focus on leveraging this \textit{editability} to generate diverse training samples---including corner cases and critical areas---to enhance downstream metrology and inspection tasks.
Simultaneously, we aim to address the observed complexity trade-off by redesigning a more token-efficient Domain-Specific Language and improving the Vision-Language Model's long-context reasoning capabilities.

\bibliographystyle{ieeetr}
\bibliography{references}

\end{document}